# PRECISE Framework: GPT-based Text For Improved Readability, Reliability, and Understandability of Radiology Reports For Patient-Centered Care

Satvik Tripathi[1,3,4], Liam Mutter[1,3], Meghana Muppuri[1,3], Suhani Dheer[2,3,4], Emiliano Garza-Frias[1,4], Komal Awan[1], Aakash Jha[5], Michael Dezube[1,4], Azadeh Tabari[1,3,4], Christopher P. Bridge[1,3,4*], Dania Daye[1,3,4,#*]

1. Department of Radiology, Massachusetts General Hospital, Boston, MA
2. Department of Psychiatry, Brigham and Women's Hospital, Boston, MA
3. Athinoula A. Martinos Center for Biomedical Imaging, Charlestown, MA
4. Harvard Medical School, Boston, MA
5. Julia R. Masterman High School, Philadelphia, PA

*co-senior authors
#Corresponding author: ddaye@mgh.harvard.edu

## Abstract

**Objective:** This study introduces and evaluates the PRECISE (Patient-Focused Radiology Reports with Enhanced Clarity and Informative Summaries for Effective Communication) framework, powered by OpenAI's GPT-4, aimed at enhancing patient understanding and engagement by providing clearer and more accessible radiology reports at the sixth-grade level.

**Design:** The PRECISE framework was assessed using 500 chest X-ray reports, employing standardized metrics such as Flesch Reading Ease, Gunning Fog Index, and Automated Readability Index to evaluate readability. Clinical volunteer assessments gauged reliability, while non-medical volunteers assessed understandability.

**Setting:** The study focused on chest X-ray reports, utilizing a diverse dataset and multiple graders to ensure comprehensive evaluation and generalizability.

**Participants:** The data utilized comprised 500 chest X-ray reports, ensuring a robust representation of medical findings.

**Interventions:** The intervention involved implementing the PRECISE framework, utilizing GPT-4 to generate patient-friendly summaries of radiology reports, and simplifying medical language.

**Main Outcome Measures:** Readability, reliability, and understandability were the primary outcome measures. Readability scores significantly improved, from an initial mean Flesch Reading Ease score of 38.28 to a mean score of 80.82 (p-value<0.001). Gunning Fog Index scores improved from an initial mean score of 13.04 to a mean score of 6.99 (p-value<0.001). ARI scores improved from an initial score of 13.33 to a mean score of 5.86 (p-value<0.001). Clinical volunteer assessments found 95% of the summaries reliable. Non-medical volunteers rated 97% of the PRECISE-generated summaries as fully understandable.

**Results:** Statistical analyses demonstrated significant differences (p-value<0.001) in readability scores between the "generated PRECISE text" and the "original report" groups, validated through t-tests, regression analyses, and Mann-Whitney U tests. The robustness of findings was underscored by consistently significant p-values across multiple methodologies.

**Conclusions:** The application of the PRECISE framework, powered by GPT-4, significantly enhances the readability and understandability of radiology reports. With improved reliability and patient-friendly summaries, this approach holds promise for fostering patient engagement and understanding in healthcare decision-making. The PRECISE framework represents a pivotal step towards more inclusive and patient-centric care delivery.

## 1. Introduction

In healthcare, a patient's capacity to make well-informed decisions through participation in their clinical care is crucial. This, however, is challenging, given the inherent complexity of medical information contained in medical documents, including radiology reports and treatment options, which often take years of specialized training to understand[1]. The imperative to convey medical information to patients in an intelligible and understandable manner is integral to patient-centric healthcare. In fact, patient-facing documents are recommended to be written at a 5th-grade level; however, they are often found to be much more complicated, deviating from the recommended standards[2]. Patients equipped with health literacy and awareness tend to engage in their own care actively and, consequently, exhibit superior health outcomes and satisfaction with the treatment provided[3,4]. Furthermore, though it is essential for physicians to offer comprehensive and detailed explanations of patients' conditions, this practice significantly contributes to the burden of administrative workload. It also imposes burnout and additional responsibilities on healthcare providers, affecting the overall quality of care[5].

Recently, AI-based conversational Large Language Models (LLMs) such as Open AI's Chat-GPT and advances in Natural Language Processing (NLP) have shown promise in solving this challenge[6,7]. These next-generation models are systems consisting of billions of parameters trained on vast datasets to produce a chat interface that can converse with informed, comprehensive, and human-like responses. Through training on huge and varied corpora of text, they possess the ability to interpret semantic complexities in prompts, allowing them to understand and explain complex topics. Despite concerns about the accuracy of the information contained within the generated text, they have demonstrated positive results in clinical applications. LLMs have proven successful in emergency departments for tasks such as referrals, structured reporting, and context-based chatbots aligned with the ACR Appropriateness Criteria. The GPT model variants have exhibited high performance, even achieving notable results on radiology board-style examinations.[8–11].

A notable instance where the confluence between medical information and patient communication occurs is in radiology reporting. Radiology reports play a crucial role in medical diagnostics, providing valuable insights and interpretations of medical imaging studies to guide patient care [12,13]. However, they often include very technical language and lack patient-centered information, which can leave patients confused about the status of their disease [14,15].

Since the implementation of the 21st Century Cures Act on April 5th, 2021, which mandates the release of all medical documents after its completion, patients' immediate access to radiological reports has presented new challenges in physician-patient communication. The exposure to documents containing highly technical medical jargon has made it difficult for patients to comprehend the reports, often leading to confusion and anxiety.[16] The RSNA-ACR Public Information Website Committee has recognized the lack of resources available to assist patients in this area. Consequently, they have developed an online guide to help patients navigate the different sections of a radiology report. With the recent advancements in large language models and their application in various medical domains, we hypothesize that implementing the PRICISE framework could effectively address the challenges posed by the jargon barrier that arose following the implementation of the 21st Century Cures Act.[17]

In this study, we explored the use of the Large Language Model GPT-4 to improve radiology reporting using the PRECISE (Patient-Focused Radiology Reports with Enhanced Clarity and Informative Summaries for Effective Communication) framework. We demonstrate that these enhanced text reports can improve medical communication with patients using principles of relevance, explicitness, conciseness, and effectiveness in a way that minimizes the administrative burden on physicians. A potential application of this process and its integration into a physician's workflow is demonstrated in Figure 1. The report is automatically translated into PRECISE text and needs only be briefly validated by the radiologist before presentation to the patient. These PRECISE texts effectively present the results of the patients' radiology reports in suitably understandable terms catering to those with limited or no medical expertise. Our results show promising potential and application for improvements in patient-centric radiology and communication by enhancing patient understanding and minimizing the burden on the physician.

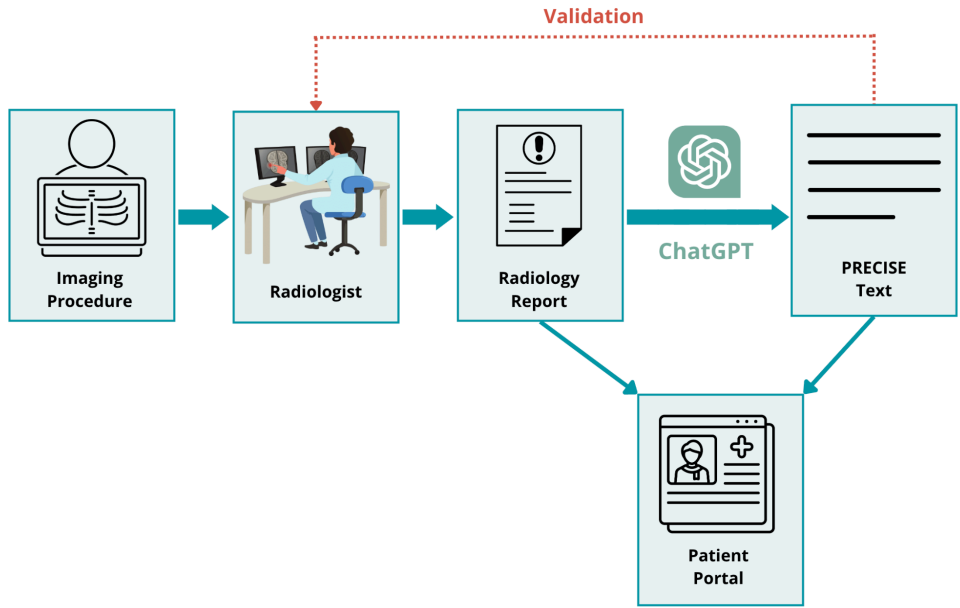

**Figure.1.** Illustration of the PRECISE framework in clinical workflow

## 2. Methods

### 2.1. Dataset

The data used in this study was sourced from the publicly available chest X-ray dataset from the University of Indiana[18]. We applied multiple inclusion and exclusion criteria to narrow down the data search to a smaller cohort. As shown in Figure 2, reports that included invalid (non alphanumeric) characters, had empty values, or contained fewer than 5 words were removed from the selected dataset. Reports containing invalid non-textual characters could not be analyzed by GPT-4 and empty reports were not useful to analyze. In addition, reports under 5 words were determined to not be useful cases as they often contained very little or very simple information. These criteria allowed us to retrieve a useful and diverse dataset to run the experiment on. We retrieved 500 X-ray reports that met all the criteria and were used for the experiments in this study. For each of the 500 reports, a corresponding PRECISE text was generated using the GPT-4 model.

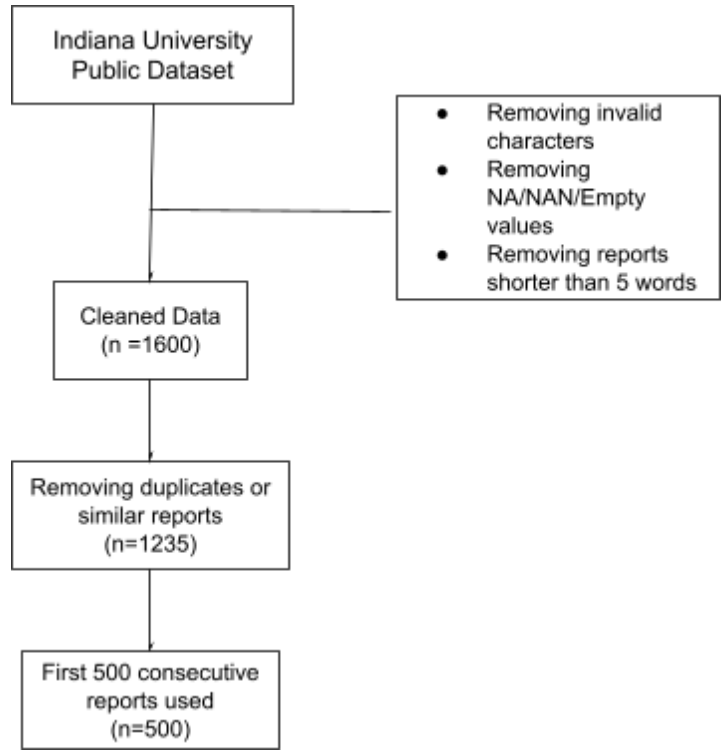

**Figure.2**. Illustration of dataset selection

## 2.2. Large Language Model

OpenAI's GPT-4 model was utilized to perform and test our hypothesis. GPT, or Generative Pre-trained Transformer, essentially based on a Decoder-only architecture for text generation. It utilizes attention mechanisms and decreases the time taken to generate results. The GPT model is trained on very large datasets of text scraped from the internet and then fine-tuned using reinforcement learning with human feedback. The current number of parameters is estimated, since most of the information is yet not public and the model might have more than a trillion[19]. This model has performed increasingly well on NLP benchmarks. With RLHF, it received a 60% on the TruthfulQA Adversarial Question benchmark, a 15 percentage point increase from GPT-3.5, which evaluated the model's ability to give factual information. In the MMLU benchmark, a benchmark designed to test a model's knowledge in 57 different subjects, it received 86.4%, again a 15 percentage point increase from its predecessor[19].

## 2.3. Prompt Engineering

Prompt engineering is a critical part of working with LLMs. The way a prompt is worded and inputted to an LLM can greatly affect the quality of the output text. Prompt engineering, in the case of GPT-4, is the process of revising the context and information in the prompt to produce a result more closely aligned with the desired output, i.e., an easier-to-read and well-translated report [21]. In our study, we utilized a given radiology report to generate a patient-friendly summary at around the sixth-grade level [22,23].

This led to the prompt: "Generate a paragraph summarizing the radiology report text at a 6th-grade level and in a patient-friendly manner." The generated response was subsequently saved alongside the original text of the report.

## 2.4. Evaluation

The generated text was evaluated for readability, reliability, and understandability. Readability measures the complexity of any given text and the reading difficulty level (i.e., sentence length, complex wording, syllables). This metric represents the difficulty level for a patient to simply read the report's words. We used three different previously validated tests. Reliability was evaluated by medical professionals with MDs. The purpose of this test was to evaluate the accuracy of the medical information generated in the PRECISE text, i.e. whether the generated result was complete and accurately represented all the information contained in the original report. Our understandability tests were designed to analyze purely whether the information given in the text was understandable. Unlike readability, this did not involve whether the language of the text was hard to read but whether the information could be parsed and understood by a non-medical audience.

### 2.4.1. Readability Tests

For this score, three metrics were used. The Flesch Reading Ease (FRE) test, Gunning Fog Index (GFI), and Automated Readability Index (ARI). These tests are standard in literature and widely used in evaluating the readability of text [24–29]. Readability can be subjective, a text that scores well in one index might score poorly in another. Using three different metrics for measuring the difficulty of readability can make general trends in readability more obvious. In addition, measuring different qualities of the text can tell us more about what exactly makes the text easy or difficult to read. For example, if a text scores better on the FRE score than on the ARI, this could tell us something about the conversationality of the text and whether the text is more conversational or technical. A summary of these tests and their qualities can be seen in Figure 3.

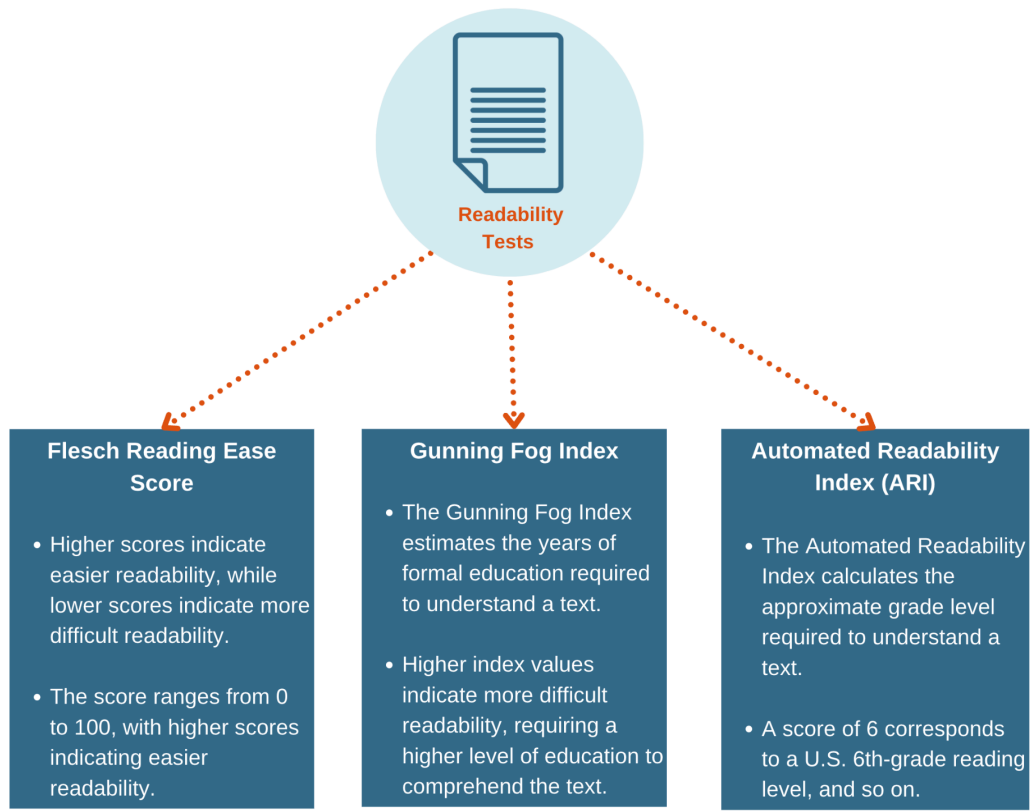

**Figure.3.** Diagram of the readability tests with descriptions of each test

**Flesch Reading Ease**

The Flesch Reading Ease (FRE) score is one test used to evaluate the readability of text. This metric measures the average sentence length and average syllable count per word. The exact formula is:

$$206.835 - 1.105\left(\frac{W}{ST}\right) - 84.6\left(\frac{W}{SY}\right) \quad \text{- (1)}$$

where W is total words, ST is total sentences, and SY is total syllables [30].

The FRE score is a useful general measure of readability in a wide variety of settings. It gives a good sense of whether a body of text contains long strings of text with some difficult language. However, not all complicated sentences involve long words with many syllables. These sentences might not receive an accurate rating from this score. The interpretation of the score is shown in Table 1 [31].

| Score | US School Level | Difficulty |
| --- | --- | --- |
| 100.0-90.0 | Fifth Grade | Easily understood by 11yo |
| 90.0–80.0 | Sixth Grade | Easy and conversational |
| 80.0-70.0 | Seventh Grade | Fairly easy |
| 70.0-60.0 | Eighth/Ninth Grade | Plain english, easy for 13-15yo |
| 60.0-50.0 | Tenth/Twelfth Grade | Fairly difficult |
| 50.0-30.0 | College | Difficult |
| 30.0-10.0 | College Graduate | Very difficult |
| 10.0-0.0 | Professional | Extremely difficult |

**Table.1.** Higher scores represent easier reading. 0-30 represents a high level of difficulty, 90-100 represents very easy reading. Scores 10 or less represent text only understandable by professionals in the field.

**Gunning Fog Index**

Secondly, we used the Gunning Fog Index. As shown in Table 2, this index represents the years of formal education required to understand a body of text [2,32]. The exact formula is

$$0.4\left[\frac{W}{S} + 100\frac{CW}{W}\right] \quad - (2)$$

where W is words, S is sentences, and CW is complex words. A complex word is a word over 3 syllables. Higher index values mean the text is harder to read, and that, comparatively, more education would be required to understand the result.

The Gunning Fog index gives information on the clarity and simplicity of a text. This index will pick up instances of complicated language with many words per sentence. However, it may overestimate the difficulty of some sentences. Simple words over three syllables are marked as complicated and increase the score when it may not be appropriate.

| Fog Index | Grade (US) Reading Level |
|---|---|
| 17 | College Graduate |
| 16 | College Senior |
| 15 | College Junior |
| 14 | College Sophomore |
| 13 | College Freshman |
| 12 | Twelfth Grade |
| 11 | Eleventh Grade |
| 10 | Tenth Grade |
| 9 | Ninth Grade |
| 8 | Eighth Grade |

| 7 | Seventh Grade |
| 6 | Sixth Grade |

**Table.2**. Higher indexes represent more difficulty. Index corresponds to US grade school level of difficulty.

**Automated Readability Index**

Lastly, we used the Automated Readability Index (ARI). This index calculates the approximate US grade level required to understand a text. The formula is

$$4.71\left(\frac{C}{W}\right) - 21.43\left(\frac{W}{S}\right) \quad \text{- (3)}$$

Where C is total characters, W is total words, and S is total sentences [33,34]. A score of "6" would correspond to a 6th-grade reading level. The ARI was developed specifically to assess the difficulty of technical writing. In cases where the text being analyzed involved technical writing, the ARI was shown to be more effective than FRE in evaluating its difficulty [35,36]. The ARI counts characters per word instead of syllables. This can often be a better indicator of difficulty than syllables in academic writing. ARI might report difficulty more inaccurately in other contexts, like conversational and public-facing information.

| Score | Age | Grade Level |
|---|---|---|
| 1 | 5-6 | Kindergarten |
| 2 | 6-7 | First/Second Grade |
| 3 | 7-9 | Third Grade |
| 4 | 9-10 | Fourth Grade |
| 5 | 10-11 | Fifth Grade |
| 6 | 11-12 | Sixth Grade |
| 7 | 12-13 | Seventh Grade |
| 8 | 13-14 | Eighth Grade |
| 9 | 14-15 | Ninth Grade |
| 10 | 15-16 | Tenth Grade |
| 11 | 16-17 | Eleventh Grade |
| 12 | 17-18 | Twelfth Grade |
| 13 | 18-24 | College Student |
| 14 | 24+ | Professor |

**Table.3.** Higher index means more difficulty. Index roughly corresponds to the US grade level of difficulty. 13 represents a college level of difficulty. 14 or more represents a professional/professor level.

### 2.4.2. Reliability Test

The texts were analyzed for medical reliability, or whether the model was providing accurate medical information summarizing exactly what was in the report correctly with no extrapolation. This scoring was done by 3 clinical volunteers with a medical degree. These 3 volunteers, who we will refer to as A, B, and C, were asked to read both the original radiology report and the corresponding generated summary and then give the summary a score from 0-2 based on the criteria shown in Figure 4. We ran a percentage agreement test on the scores from these volunteers to determine whether this metric was valid and the scores were accurate. There was an 86.17% agreement between A and B, a 94.78% agreement between B and C, and 85.57% agreement between A and C.

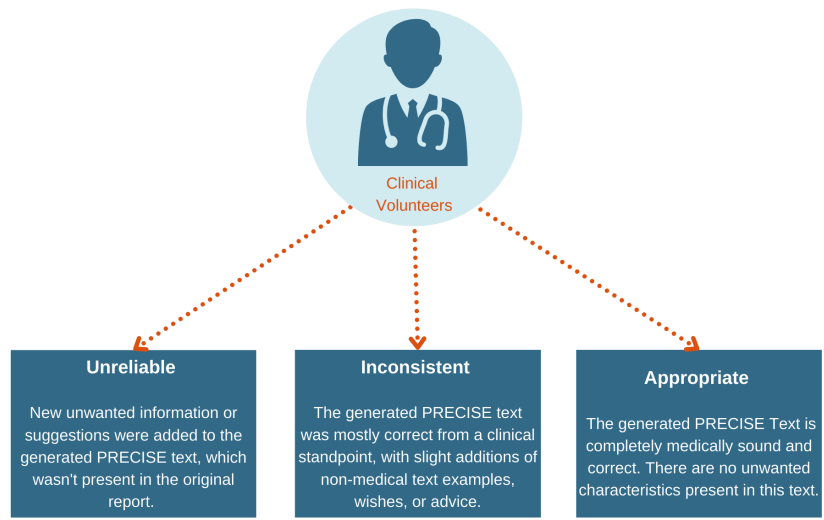

**Figure 4.** Illustration of reliability testing by clinical volunteers with the description of each grade. Grading of 0, 1, and 2 is Unreliable, Inconsistent, and Appropriate respectively.

### 2.4.3. Understandability Test

The volunteers scored a total of 1000 texts, including both the original reports and the generated summary. Volunteers were blinded to whether each text was original or generated, and the texts were presented to the readers in a random order, shuffling the two sets of texts together. This score was generated by 4 non-medical US-based volunteers: 2 high-school students, and 2 college students. Each volunteer was given only one text at a time and asked to score it from 0-2 based on the criteria shown in Figure 5. A Mann Whitney U was performed with the null hypothesis that generated reports had similar understandability scores to the original reports. In addition, we calculated a Cohen's kappa score to determine the agreement between graders on the score of each text to prove that our scoring system was valid and consistently represented the understandability of the text.

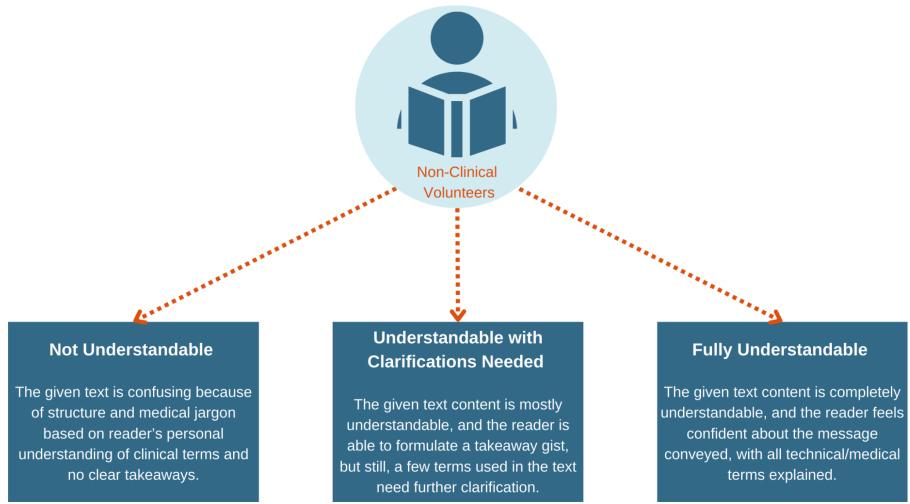

**Figure 5.** Illustration of understandability testing with descriptions of grades given by non-clinical volunteers

The results of each test were only calculated once all scoring was done, as such the results of each test were kept private from the graders. Each participant had the same instructions and used the same method of grading and analysis for all data and there was no leakage or bias in the review process.

## 3. Results

### 3.1 Readability scores

The readability scores obtained from the study indicate that the PRECISE framework significantly simplifies medical text. With the Flesch Reading Ease (FES) score averaging at 81±3.9, the majority of the texts are of lower difficulty, being "easy and conversational" and equivalent to a sixth-grade reading level. The Gunning Fog Index score averaging at 7±1.9 with a positively skewed distribution of scores reflects that the majority of the LLM outputs are readable by an average seventh grader. The Automated Readability Index (ARI) score averaging at 6±2.8 indicates that the LLM outputs have a readability of a 6th-grade level or higher. These scores collectively reinforce the PRECISE framework's capability to convert medical findings into an easily readable format for the general public (Figure. 6).

## A) Radiology Report Text

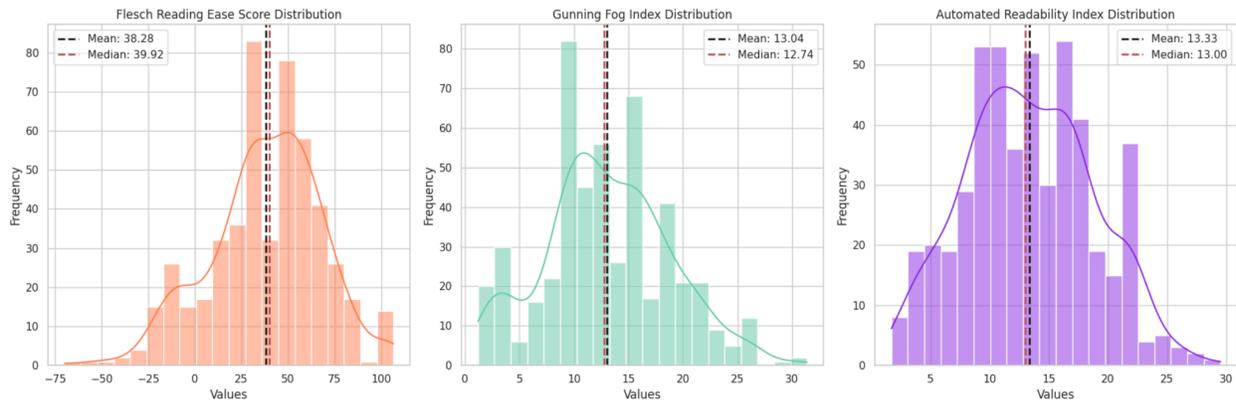

## B) Generated PRECISE Text

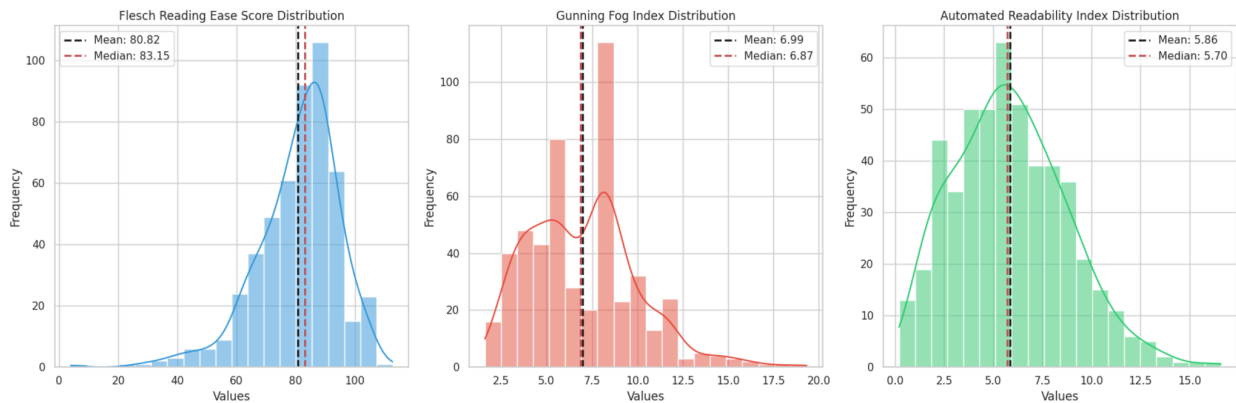

**Figure 6.** The data's frequency distribution of readability scores for the original radiology report and the generated PRECISE text based on three common metrics: Flesch Reading Ease, Gunning Fog Index, Automated Reliability Index.

In this analysis, we employed various statistical methods to compare readability scores between two groups denoted as "generated PRECISE text" and "original report." Initially, we conducted t-tests and regression analyses for three readability metrics to evaluate differences in means and relationships between the groups. Subsequently, we extended our analysis to include non-parametric Mann-Whitney U tests, suitable for non-normally distributed data. The p-values obtained from these tests were incorporated into a heatmap, providing a comprehensive visual representation of statistical comparisons (Figure 7). This multi-methodological approach allowed us to assess group differences while considering both parametric and non-parametric aspects, enhancing the robustness of our findings and accommodating potential violations of assumptions.

The results of our statistical analyses reveal compelling evidence of significant differences (p-value<0.001). The t-test results indicate extremely low p-values for all three readability metrics, suggesting that the means of the two groups are significantly distinct. Similarly, the regression analyses reinforce these findings, with p-values well below the conventional significance threshold. The Mann-Whitney U tests, which are robust to non-normal distributions, consistently show remarkably low p-values, further supporting the conclusion of significant differences between the groups. These results collectively underscore the robustness of our findings across multiple statistical methodologies, enhancing the reliability and validity of our conclusions regarding the distinctiveness of readability scores in groups "generated PRECISE text" and "original report."

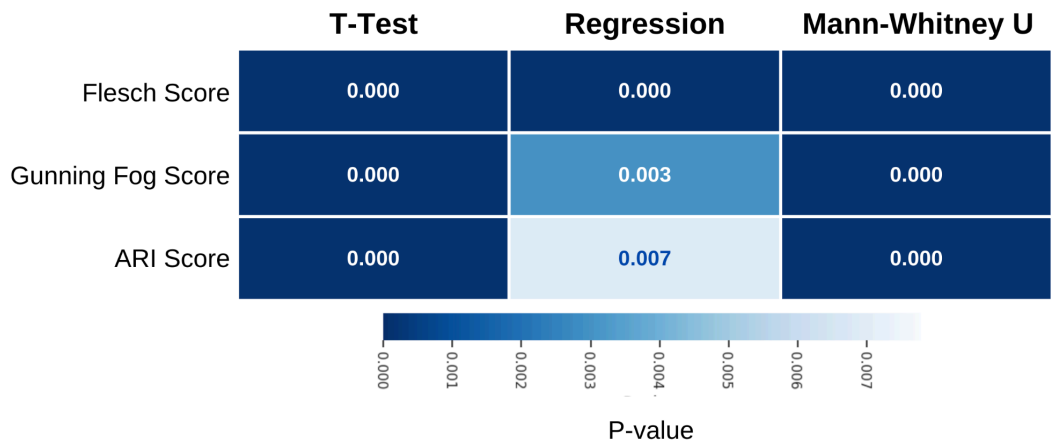

**Figure 7.** Heatmap illustrating the p-values from t-tests, regression analyses, and Mann-Whitney U tests for three readability metrics comparing groups "generated PRECISE text" and "original report." The color intensity represents the statistical significance of differences, providing a comprehensive visualization of the multi-methodological approach used to assess the distinctiveness of readability scores between the two groups.

### 3.2 Reliability scores

The reliability test of the PRECISE framework indicates a strong consensus among medical professional graders regarding the accuracy of the generated summaries as shown in Figure 8. With 95% of the texts being classified as "appropriate", this reflects a high degree of confidence in the clinical validity of the data presented. The small percentages of 4% categorized as "inappropriate" and 1% as "unreliable" suggest that there are occasional deviations from optimal translation of the medical data.

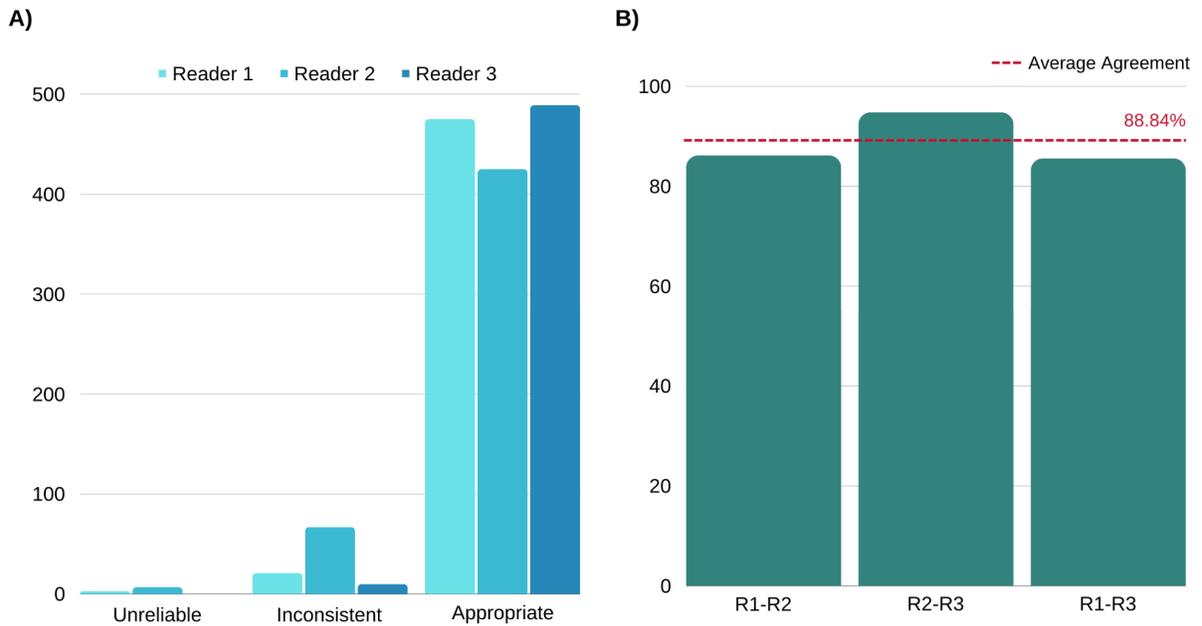

**Figure 8.** (A) The stacked bar chart on the left shows the distribution of reliability scores of the PRECISE text summaries rated by medical professionals. (B) The percentage agreement between raters is shown in the adjoining bar chart on the right.

### 3.3 Understandability scores

For the understandability test, blind test volunteers, 97% of the time, chose "not-understandable" for the actual report and "fully understandable" for the PRECISE text. This contrast between the original radiology reports and the PRECISE-generated texts is graphically illustrated in Figure. 9. Additionally, the consistency across different evaluators suggests that the PRECISE framework helped simplify medical language for the generated text's understandability scores.

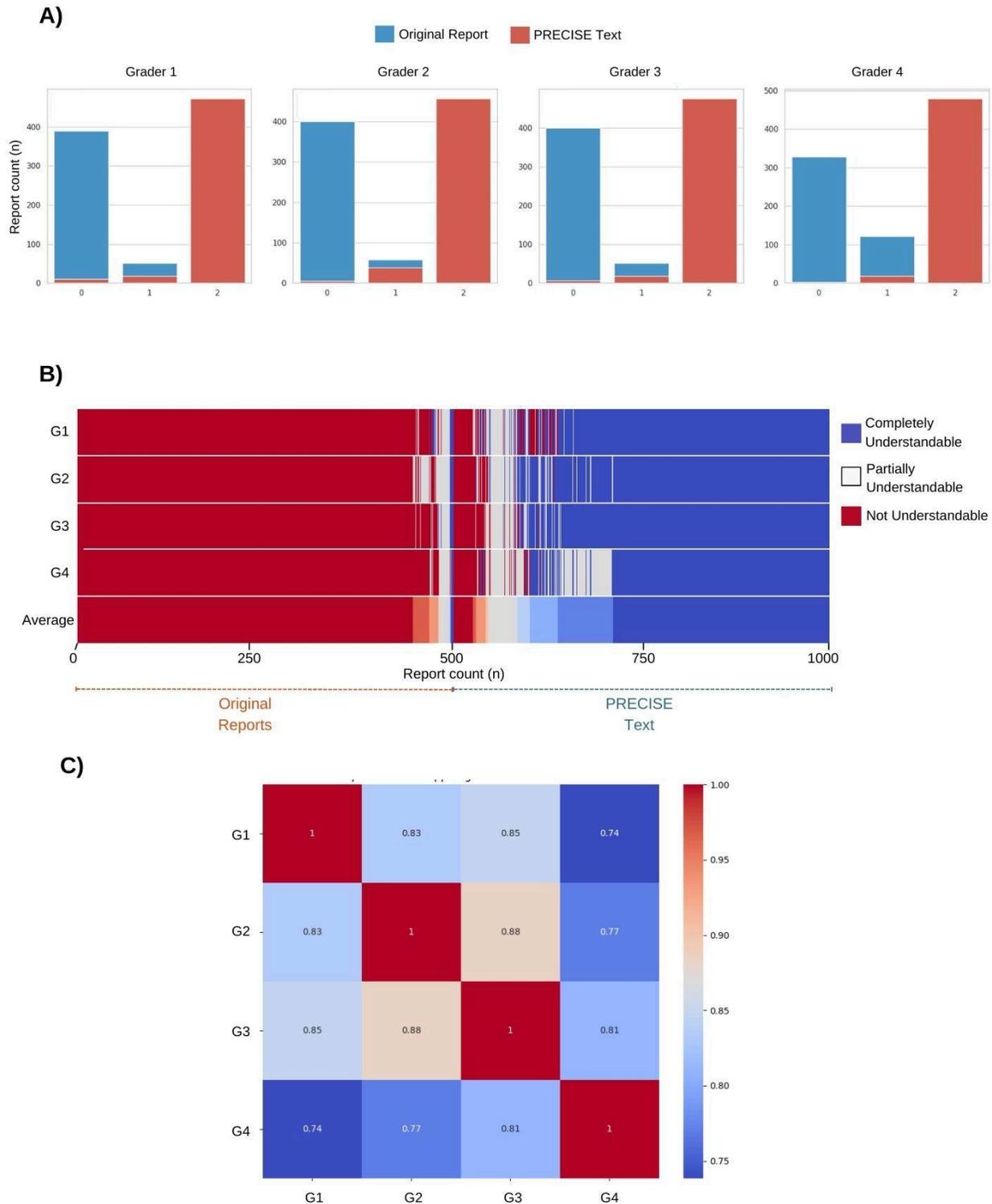

**Figure 9.** A) Individual bar charts for each grader showing preference in understandability of medical texts translated using the PRECISE framework. B) Heatmap indicating reader agreement and understandability between the original reports and PRECISE text. Each vertical bar represents the understandability grade of a single report for each of the four graders and the average grade. Within the original reports (left half) and PRECISE texts (right half), reports are ordered by average grade C) Heatmap of Cohen's Kappa showing a high inter-rater consistency, for each grader represented as G1, G2, G3, and G4, respectively.

## 4. Discussion

This study is the first to investigate and determine the efficacy of the application of ChatGPT's GPT-4 in generating diagnostic reports using the PRECISE framework. This framework enhances the efficiency and clarity of the radiology reports, making them more understandable to patients. Patients typically encounter challenges in understanding radiology reports that use complex medical language, compounded by the lack of image visualization[38]. Improved health literacy is essential for improved medical outcomes and patient satisfaction. Our method centered on using GPT-4 to auto-generate these patient-friendly summaries, eliminating the traditional challenges associated with complex medical language.

Patients universally encounter challenges in interpreting radiology reports and understanding their significance in diagnosis and clinical management. These reports largely utilize medical terminology tailored to healthcare professionals rather than the patients themselves[23,39]. While this specialized language is essential to clinical accuracy, it poses a barrier to patients' understanding. This not only impairs their comprehension but also contributes to misinterpretation and anxiety[23,40,41]. The PRECISE framework addresses this imbalance by re-envisioning radiology reports with an emphasis on patient-centered information without placing further burden on radiologists. Our framework streamlines medical content into clinically accurate summaries that are also accessible and easily understandable. Through enhanced patient comprehension fostered through the PRECISE framework, individuals are better informed about their health without compromising the clinical integrity of the report. This highlights the role of our PRECISE framework in patient-centric medical reporting.

The application of three readability metrics to our PRECISE framework has been investigated in our study in an effort to make radiology reports more readable to patients. The Flesch Reading Ease metric applied to our PRECISE text evaluates the average sentence length and syllable count per word. The PRECISE framework demonstrates superiority in readability in comparison to traditional radiology reports that often require higher education levels for better comprehension[23]. Our quantitative results together highlight PRECISE framework's capability to produce more comprehensible summaries of the impressions generated on radiology reports, hence reaching a broader audience. With further research and application, our standardized approach has the potential to incorporate more patient engagement and encourage the active role of patients in clinical decision-making while relieving physicians of administrative load.

Most of the generated summaries were classified as "appropriate," indicating a significant degree of medical accuracy to the original reports. This positive classification highlights the scope of the PRECISE framework to facilitate the generation of patient-friendly yet reliable and accurate reports without medical jargon. This addresses a crucial aspect of medical communication, eliminating misinterpretations by the patients. This patient-focused approach bridges this identified challenge without compromising the level of medical integrity in the summaries [42]. None of the generated summaries corresponded to unreliable summaries, further supporting the dependability of the PRECISE framework in ensuring the delivery of medical information to patients.

In further evaluation of our PRECISE framework, much of our focus was placed on its ability to enhance the understandability of medical texts, specifically non-medical readers. Most of the texts were rated as being "fully explainable." This demonstrates significant promise in improving comprehension for patients. These collective challenges associated with understanding radiology reports can affect patient participation in clinical decision-making, leading to non-compliance to therapeutic interventions. In contrast, our PRECISE-generated texts reflect a reduction in these challenges through the facilitation of clearer and more accessible information for those reading them. Thus highlighting the potential of the PRECISE framework in improving patient-centric communication in healthcare.

The PRECISE framework that our study utilizes integrates novel components through the incorporation of ChatGPT to optimize the readability and understandability of radiology reports. This method has demonstrated, through quantifiable metrics discussed previously, an enhanced level of dependability in delivering medical information. Through the utilization of ChatGPT, medical jargon is accurately translated into accessible language without compromising the integrity of the medical information. In comparison to existing methods of radiology reporting, the PRECISE framework reinforces the readability and understandability while addressing the gist of the diagnostic report.

Our study identifies the PRECISE framework as an innovative tool in patient-centric communication and medical reporting. When medical jargon is translated into patient-friendly language without compromising the integrity of the generated text, patients are more likely to be encouraged to participate in clinical decision-making along with their healthcare providers actively. This translates into patient satisfaction and involvement due to the inclusive and accessible nature of the language used in the PRECISE texts. Hence, an informed patient is more likely to clearly understand their diagnosis and related management plan, while building trust with their healthcare provider, contributing to more positive health outcomes through active engagement in their own healthcare. Our methods and results are highly reproducible as several volunteers contributed to the data validation.

The utilization of GPT-4 within the PRECISE framework emerges as a key driver of these positive outcomes, leveraging the model's capabilities in generating patient-friendly summaries at an eighth-grade reading level. The prompt engineering process, guided by a patient-centric approach, aligns with the framework's overarching goal of facilitating transparent and comprehensible communication. While the study demonstrates promising results, it is imperative to acknowledge certain limitations, such as the use of a single dataset from the University of Indiana and the inclusion of 500 reports from a specific institution. Future research should explore the framework's applicability across diverse datasets, institutions, and diagnostic imaging modalities, ensuring a more comprehensive validation of its dependability and generalizability.

The results of this study substantiate the transformative potential of the PRECISE framework in addressing longstanding challenges associated with the readability, reliability, and understandability of radiology reports. The framework's success in simplifying medical text, as evidenced by favorable readability scores, positions it as a valuable tool for enhancing communication between healthcare providers and patients. The high-reliability scores affirm the clinical validity of the generated summaries, instilling confidence in the framework's accuracy in conveying essential medical information. There is a valuable opportunity for the integration of the PRECISE-generated summaries of radiology reports into the standard radiology reporting templates. Not to mention, the applicability of the PRECISE framework extends outside of medical reporting and into the broader aspect of patient-centric communication.

## 5. Conclusion

In summary, our investigation underscores the transformative potential of the PRECISE framework in reshaping the landscape of radiology reporting through its steadfast commitment to a patient-centered paradigm. The empirical validation of the framework's efficacy, as exemplified by the positive outcomes of the explainability test, substantiates its role in augmenting comprehension among audiences outside the medical domain. The consequential implications of this innovative approach manifest in its ability to empower patients with a nuanced comprehension of their medical conditions. The provision of accessible and comprehensible radiology reports by the PRECISE framework serves as a catalyst for fostering meaningful and informed dialogues between patients and healthcare providers. This newfound clarity propels patients towards active engagement in their healthcare, thereby cultivating a collaborative ecosystem where patients assume an integral role in the decision-making processes. In essence, the PRECISE framework not only optimizes communication between medical professionals and patients but also fortifies the patient-provider relationship, heralding a healthcare milieu characterized by transparency and patient-centricity. This paradigm shift holds the promise of significantly enhancing the overall quality and efficacy of healthcare delivery.

**Author Contributions**
The conceptualization of this study was a collaborative effort involving S.T., D.D., and C.P.B. Methodology design and development were carried out by S.T., D.D., and C.P.B. The investigation phase included contributions from S.T., L.M., M.M., S.D., E.G.F., K.A., M.D., A.T., D.D., and C.P.B. The writing of the original draft was a collective effort led by S.T., L.M., M.M., S.D., and E.G.F. Subsequent review and editing were conducted by S.D., A.T., M.D., C.P.B., and D.D. Visualization elements were contributed by S.T. The supervision of the entire project was overseen by S.T., C.P.B., and D.D. All authors have thoroughly reviewed and provided their consent to the published version of the manuscript.

**Funding**
This research received no external funding.

**Conflicts of Interest**
The authors declare no conflict of interest.

**References**

1. Fossa AJ, Bell SK, DesRoches C. OpenNotes and shared decision making: a growing practice in clinical transparency and how it can support patient-centered care. *J Am Med Inform Assoc*. 2018;25(9):1153-1159.

2. Hansberry DR, John A, John E, Agarwal N, Gonzales SF, Baker SR. A critical review of the

readability of online patient education resources from RadiologyInfo.Org. *AJR Am J Roentgenol*. 2014;202(3):566-575.

3.  Swanwick T, Forrest K, O'Brien BC. *Understanding Medical Education: Evidence, Theory, and Practice*. John Wiley & Sons; 2018.

4.  Densen P. Challenges and opportunities facing medical education. *Trans Am Clin Climatol Assoc*. 2011;122:48-58.

5.  Heuer AJ. More Evidence That the Healthcare Administrative Burden Is Real, Widespread and Has Serious Consequences Comment on "Perceived Burden Due to Registrations for Quality Monitoring and Improvement in Hospitals: A Mixed Methods Study." *Int J Health Policy Manag*. 2022;11(4):536-538.

6.  Homolak J. Opportunities and risks of ChatGPT in medicine, science, and academic publishing: a modern Promethean dilemma. *Croat Med J*. 2023;64(1):1-3.

7.  Tripathi S, Sukumaran R, Cook TS. Efficient healthcare with large language models: optimizing clinical workflow and enhancing patient care. *J Am Med Inform Assoc*. Published online January 25, 2024. doi:10.1093/jamia/ocad258

8.  Barash Y, Klang E, Konen E, Sorin V. ChatGPT-4 Assistance in Optimizing Emergency Department Radiology Referrals and Imaging Selection. *J Am Coll Radiol*. Published online July 7, 2023. doi:10.1016/j.jacr.2023.06.009

9.  Adams LC, Truhn D, Busch F, et al. Leveraging GPT-4 for Post Hoc Transformation of Free-text Radiology Reports into Structured Reporting: A Multilingual Feasibility Study. *Radiology*. 2023;307(4):e230725.

10. Rau A, Rau S, Zoeller D, et al. A Context-based Chatbot Surpasses Trained Radiologists and Generic ChatGPT in Following the ACR Appropriateness Guidelines. *Radiology*. 2023;308(1):e230970.

11. Bhayana R, Krishna S, Bleakney RR. Performance of ChatGPT on a Radiology Board-style Examination: Insights into Current Strengths and Limitations. *Radiology*. 2023;307(5):e230582.

12. Kemp J, Short R, Bryant S, Sample L, Befera N. Patient-Friendly Radiology Reporting-Implementation and Outcomes. *J Am Coll Radiol*. 2022;19(2 Pt B):377-383.

13. Demystifying Radiology Reports. Accessed February 3, 2024. https://www.acr.org/Practice-Management-Quality-Informatics/ACR-Bulletin/Articles/December-2021/Demystifying-Radiology-Reports

14. Farmer CI, Bourne AM, O'Connor D, Jarvik JG, Buchbinder R. Enhancing clinician and patient understanding of radiology reports: a scoping review of international guidelines. *Insights Imaging*. 2020;11(1):62.

15. Mezrich JL, Jin G, Lye C, Yousman L, Forman HP. Patient Electronic Access to Final Radiology Reports: What Is the Current Standard of Practice, and Is an Embargo Period Appropriate? *Radiology*. 2021;300(1):187-189.

16. Bruno B, Steele S, Carbone J, Schneider K, Posk L, Rose SL. Informed or anxious: patient preferences for release of test results of increasing sensitivity on electronic patient portals. *Health Technol* . 2022;12(1):59-67.

17. Website. Demystifying radiology reports. Demystifying Radiology Reports | American College of Radiology. (n.d.). https://www.acr.org/Practice-Management-Quality-Informatics/ACR-Bulletin/Articles/December-2021/Demystifying-Radiology-Reports Accessed 1/29/2024

18. Demner-Fushman D, Kohli MD, Rosenman MB, et al. Preparing a collection of radiology examinations for distribution and retrieval. *J Am Med Inform Assoc*. 2016;23(2):304-310.

19. Baktash JA, Dawodi M. Gpt-4: A Review on Advancements and Opportunities in Natural Language Processing. *arXiv [csCL]*. Published online May 4, 2023. http://arxiv.org/abs/2305.03195

20. OpenAI, :, Achiam J, et al. GPT-4 Technical Report. *arXiv [csCL]*. Published online March 15, 2023. http://arxiv.org/abs/2303.08774

21. Liu P, Yuan W, Fu J, Jiang Z, Hayashi H, Neubig G. Pre-train, Prompt, and Predict: A Systematic Survey of Prompting Methods in Natural Language Processing. *arXiv [csCL]*. Published online July 28, 2021. http://arxiv.org/abs/2107.13586

22. Patil S, Yacoub JH, Geng X, Ascher SM, Filice RW. Radiology Reporting in the Era of Patient-Centered Care: How Can We Improve Readability? *J Digit Imaging*. 2021;34(2):367-373.

23. Readability of radiology reports: implications for patient-centered care. *Clin Imaging*. 2019;54:116-120.

24. Si L, Callan J. A statistical model for scientific readability. In: *Proceedings of the Tenth International Conference on Information and Knowledge Management*. CIKM '01. Association for Computing Machinery; 2001:574-576.

25. Kouamé JB. Using readability tests to improve the accuracy of evaluation documents intended for low-literate participants. Accessed January 22, 2024. https://journals.sfu.ca/jmde/index.php/jmde_1/article/download/280/283/0

26. Rahmawati LE, Sulistyono Y. Assessment and evaluation on text readability in reading test instrument development for BIPA-1 to BIPA-3 / Laili Etika Rahmawati and Yunus Sulistyono. *Asian Journal of University Education (AJUE)*. 2021;17(3):51-57.

27. Onder CE, Koc G, Gokbulut P, Taskaldiran I, Kuskonmaz SM. Evaluation of the reliability and readability of ChatGPT-4 responses regarding hypothyroidism during pregnancy. *Sci Rep*. 2024;14(1):243.

28. Ayers JW, Zhu Z, Poliak A, et al. Evaluating Artificial Intelligence Responses to Public Health Questions. *JAMA Netw Open*. 2023;6(6):e2317517.

29. Moons P, Van Bulck L. Using ChatGPT and Google Bard to improve the readability of written patient information: A proof-of-concept. *Eur J Cardiovasc Nurs*. Published online August 21, 2023. doi:10.1093/eurjcn/zvad087

30. Wrigley Kelly NE, Murray KE, McCarthy C, O'Shea DB. An objective analysis of quality and readability of online information on COVID-19. *Health Technol* . 2021;11(5):1093-1099.

31. Ferguson C, Merga M, Winn S. Communications in the time of a pandemic: the readability of documents for public consumption. *Aust N Z J Public Health*. 2021;45(2):116-121.

32. Świeczkowski D, Kułacz S. The use of the Gunning Fog Index to evaluate the readability of Polish and English drug leaflets in the context of Health Literacy challenges in Medical Linguistics: An exploratory study. *Cardiol J*. 2021;28(4):627-631.

33. Smith EA, Senter RJ. Automated readability index. *AMRL TR*. Published online May 1967:1-14.

34. Chall JS. *Readability: An Appraisal of Research and Application*. Literary Licensing, LLC; 2012.

35. Peter Kincaid J, Fishburne RP Jr, Rogers RL, Chissom BS. Derivation Of New Readability Formulas (Automated Readability Index, Fog Count And Flesch Reading Ease Formula) For Navy Enlisted Personnel. Published online 1975. Accessed January 21, 2024. https://stars.library.ucf.edu/istlibrary/56/?utm_sourc

36. Pawłowski A, Mačutek J, Embleton S, Mikros G. *Language and Text: Data, Models, Information and Applications*. John Benjamins Publishing Company; 2021.

37. Elkassem AA, Smith AD. Potential Use Cases for ChatGPT in Radiology Reporting. *American Journal of Roentgenology*. Published online April 19, 2023. doi:10.2214/AJR.23.29198

38. Alarifi M, Patrick T, Jabour A, Wu M, Luo J. Understanding patient needs and gaps in radiology reports through online discussion forum analysis. *Insights Imaging*. 2021;12. doi:10.1186/s13244-020-00930-2

39. Olthof AW, de Groot JC, Zorgdrager AN, Callenbach PMC, van Ooijen PMA. Perception of radiology reporting efficacy by neurologists in general and university hospitals. *Clin Radiol*. 2018;73(7). doi:10.1016/j.crad.2018.03.001

40. Johnson AJ, Frankel RM, Williams LS, Glover S, Easterling D. Patient access to radiology reports: what do physicians think? *J Am Coll Radiol*. 2010;7(4). doi:10.1016/j.jacr.2009.10.011

41. Rosenkrantz AB. Differences in Perceptions Among Radiologists, Referring Physicians, and Patients Regarding Language for Incidental Findings Reporting. *AJR Am J Roentgenol*. 2017;208(1). doi:10.2214/AJR.16.16633

42. Alarifi M, Patrick T, Jabour A, Wu M, Luo J. Designing a Consumer-Friendly Radiology Report using a Patient-Centered Approach. *J Digit Imaging*. 2021;34(3):705.

43. Tripathi S, Gabriel K, Dheer S, et al. Understanding Biases and Disparities in Radiology AI Datasets: A Review. *J Am Coll Radiol*. 2023;20(9):836-841.